\RequirePackage{fix-cm}
\documentclass[smallextended]{svjour3}          
\smartqed 
\usepackage{graphicx}
\usepackage{mathptmx}
\usepackage[utf8]{inputenc}
\usepackage{booktabs}
\usepackage{graphicx}
\usepackage{rotating}
\usepackage{amsmath}
\usepackage{adjustbox}
\usepackage{tikz}
\usetikzlibrary{arrows.meta, chains}
\usepackage{lineno}

\begin{document}

\title{Predicting special care during the COVID-19 pandemic: A machine learning approach}

\author{Vitor P. Bezzan \and Cleber D. Rocco}

\institute{Vitor P. Bezzan \at
              Instituto de Matemática, Estatistica e Computação Científica - Universidade Estadual de Campinas \\
              \email{vitor@bezzan.com}           
           \and
           Cleber D. Rocco \at
              Faculdade de Ciências Aplicadas - Universidade Estadual de Campinas\\
              \email{cdrocco@unicamp.br}
}

\date{Received: date / Accepted: date}

\maketitle


\begin{abstract}
More than ever COVID-19 is putting pressure on health systems all around the world, especially in Brazil. In this study we propose an analytical approach based on statistics and machine learning that uses lab exam data coming from patients to predict whether patients are going to require special care (hospitalisation in regular or special-care units). We also predict the number of days the patients will stay under such care. The two-step procedure developed uses Bayesian Optimisation to select the best model among several candidates leads us to final models that achieve 0.94 area under ROC curve performance for the first target and 1.87 root mean squared error for the second target (which is a 77\% improvement over the mean baseline), making our model ready to be deployed as a decision system that could be available for everyone interested. The analytical approach can be used in other diseases and can help the planning hospital capacity.

\end{abstract}

\section{Introduction}

The COVID-19 pandemic is showing itself as a huge challenge for Brazil and several other countries all around the world because the disease is putting tremendous pressure on public health care services. Several independent reports exist indicating very high occupation rate in intensive care units with facilities to support patients with serious respiratory tract failure and similar conditions, creating an unique opportunity to solve this problem with scientific rigour.


As the disease spreads quickly and measures of social distancing are being de-phased in several countries despite recommendations on the contrary issued by WHO and CDC, pointed by \cite{virusR0}. A tool to detect the probability to a given patient having to receive special care in hospitals using the most easily available information is more than necessary to adjust efforts in way to satisfy the needs of both hospitals and patients alike.

As pointed by \cite{collab}, the huge amount of data being captured from several sources should be put into good use for intensive training of machine learning algorithms to understand better the disease, its patients and possible prognosis, enabling informed decision-taking. To accomplish that, here we propose an analytical approach based on machine learning (ML) that uses laboratory tests data that estimates the aforementioned probability and also estimates the number of days a given patient will be under such care. For that we use widely available techniques and methods to create the basis of a decision system that can be used by anyone interested in replicating and estimating such outcomes, expanding the system to deal with other diseases whether needed.

We used data available through \cite{data}, which condensates laboratory exam data from Sírio Libanês Hospital, Albert Einstein Israeli Hospital and Fleury Laboratories (all located in city of São Paulo, Brazil). This data comprises several different types of laboratory exams performed at patients (most of the exams being blood tests). This preference for blood tests is not accidental: most of them are well-standardised and are usually cheap to perform, being accessible in most situations even for developing countries.

This article is organised as follows: in Section \ref{sec:review}, we examine some of the most relevant literature present on machine learning with a healthcare perspective; After that, in Section \ref{sec:method}, we present our analytical approach used to create ML models to predict special care probability and extend the same approaches to predict how many days any given patient will spend under such care - focusing on the overall applicability and explainability of the models trained; We then present the overall numerical results for both targets in Section \ref{sec:numerical_results}, considering the candidate models and the final selected optimised ones; At the end, we present our conclusions, limitations and possible extensions that should follow for other diseases and situations where our approach could be useful.

\section{Literature Review}
\label{sec:review}

This literature review will be focused in putting into light recent efforts on the use of ML and decision systems with a healthcare perspective. We will dive on the specific subject of COVID-19 for some of the references, but we will also give some focus to new, interesting or emerging applications for other diseases and situations to clarify the research panorama in the subject and to compare this article with others on the same field.

The use of statistical methods in healthcare for a great number of individuals comprising a great amount of data points dates back to the 50s, when the Framingham Heart Study was first established showing correlations between health measurements made by doctors (including some lab test results) and heart diseases, diabetes and obesity. See \cite{framingham} for a historical perspective and \cite{bertsimas} for a statistical point of view. This study in particular is noted as one of the finest and earliest examples of how statistics and decision systems could be implemented to help governments and policymakers to make well informed decisions that have huge impact on the quality of life and overall survival rate of an specific individual.

After the 50s, with the advent of faster computers with high-level programming languages and frameworks, several studies arose under the ML and decision systems umbrella. From medicine to economics and social sciences, these studies helped people and governments to make more scientifically informed decisions with \textit{realy huge} and diverse data coming from different sources. From now on, we will focus on recent developments.

Recent examples of ML being used to detect and diagnose different types of diseases using exam data are appearing in different contexts; In \cite{hemato}, we see classifiers being applied to detect haematological diseases sometimes better than haematologists itself, a frontier algorithms in general are reaching with substantial implications.

In \cite{blood} the authors use laboratory data on patients also to detect blood diseases. In their approach, they select several candidate models within minimal pre-treatment of data to understand which algorithm behaves better; In the present study we expand that proposing with an second optimisation procedure on the selected algorithm type to improve the specificity-sensitivity characteristics of the final optimised model. See Figure \ref{fig:steps} as scheme for better understanding.

Blood test data are also being used to detect more complex types of diseases. There is an special interest in several areas, with \cite{oncology} being one remarkable example. Their aim is to detect more than 50 types of different cancers by analysing different signatures in the DNA, showing an 99.3\% specificity rate. This article can be seen as an improvement in the field of "liquid biopsies", reducing the need of patients to undergo complicated procedures to get their diagnostic.

There is other diseases where ML algorithms-aided diagnosis could play a significant role. For example, \cite{fattyliver} applies random forests for the final selected model to predict fatty liver disease and create an indicator to separate high-risk patients from low-risk ones, effectively allowing customisation in treatments and improving overall outcomes. Stepping a little bit outside our scope, there is also a substantial number of studies using algorithms that do not rely on laboratory data to predict outcomes (examples being the use of deep learning to learn from medical images). A good review on this topic is provided by \cite{survey}, where applications for heart disease, dengue fever, hepatitis and diabetes are explored.

Looking at the interface in decision systems we can cite \cite{bacteremia} as an application of ML-backed classifiers to understand the potential of bacterial infection in a given patient on a hospital setting. Special attention is given to prioritise hospital resources and early detection of bacteremia, an infectious disease caused by microorganisms that propagate much like COVID-19. On the same topic, we can also cite \cite{ebola}, an article showing the creation of a decision system given to hospitals to predict the outcomes of Ebola in West African patients (Ebola being a highly contagious virus that demands special care of patients, in resemblance with COVID-19).

Books are also being dedicated to these subjects. In \cite{decision}, we see several applications of ML in different areas spanning disease diagnostics with laboratory data, image recognition methods, unsupervised learning and Internet of Things.

The interest in these topics are getting stronger as time passes and technology advances; Conferences and meetings are being done in several places. One interesting example is the \textit{Machine Learning for Healthcare}  \cite{MLHC} conference which took place virtually in 2020 due to COVID-19 pandemic.

Tying us specifically to COVID-19, there are several reports on the use of ML to detect the disease using laboratory data; In \cite{covid1}, the authors trained classifiers that attained an 82\%-86\% accuracy while keeping high levels of specificity and sensitivity, therefore increasing the general applicability of the method selected. There is also an example in \cite{ibero} of deep learning based methods being used to estimate the overall epidemiological parameters for the disease considering stacked Long Short-term Memory (LSTM) models and polynomial neural networks.

Some novel and interesting approaches are emerging from the necessity to diagnose patients using any data available. In \cite{newone}, we see a novel feature generation approach in X-ray images combined with optimisation techniques and high-performance computing being used to create a classifier for patients with 96-98\% accuracy. On a even more exotic front, text data is being used to diagnose patients in \cite{text}.

Considering that COVID-19 is itself a relatively novel subject, extensive reviews for articles relating it with ML algorithms are only beginning to emerge. In \cite{soliton}, we see one of the first examples.

There are two main differences between this article and the ones cited earlier. The first one is target itself: instead of predicting the presence/absence of COVID-19 in one give patient, we try to explore the probability of this patient requiring special care at the hospital (and the number of days required under special care). The second main difference is the number of algorithms: instead of focusing in one or two algorithms, we firstly considered several, and then we select the best algorithm class overall to perform the Bayesian Optimisation. Table \ref{tab:articles} summarises the findings in this section and positions our study among them.

\begin{center}
\begin{table}[ht!]
\caption{Review of machine learning for disease prediction.} 
\label{tab:articles}
\centering
\begin{tabular}{l l l}
            \hline
    		Reference & Algorithm & Key Results \\ \hline
    		\cite{bertsimas} & Logistic Regression, Random Forests & $0.72$ AUC  \\
    		\cite{ebola} & Model Ensembles & $0.80$ AUC \\
    		\cite{fattyliver} & Random Forests & $0.92$ AUC \\
    		\cite{bacteremia} & Random Forests & $0.82$ AUC \\
    		\cite{blood} & Several & $0.69$ - $0.97$ AUC \\
    		\cite{hemato} & Random Forests & $59\%$ - $80\%$ Precision \\
    		\cite{oncology} & Several & $99.3\%$ Specificity \\
    		\cite{covid1} & Random Forests, SVM and others & $92\%$ - $95\%$ Sensitivity \\
    		\cite{ibero} & LSTM & $62\%$ - $87\%$ Accuracy \\
    		\cite{newone} & DNNs & $96\%$ - $98\%$ Accuracy \\
    		\cite{text} & Naïve Bayes & $96.20\%$ Accuracy \\
    		\cite{decision} & Several & - \\
    		\cite{soliton} & Several & - \\
    		\cite{survey} & Several & - \\
    		\textbf{This article} & xgBoost + Bayesian Optimization & $0.94$ AUC \\ \hline
\end{tabular}
\end{table}
\end{center}

\section{Method}
\label{sec:method}

This section is dedicated to lay all the groundwork used in this study. Firstly we present some medical basis, showing some results and references linking blood test results and its respective impacts in COVID-19 patients. We also show the algorithmic reasoning behind all techniques involved and why we selected them.

\subsection{Medical Basis}
\label{medical_basis}

As COVID-19 being a virus is coherent to assume its infection can cause changes on patient blood tests. The article \cite{giuseppe202laboratory} brings an structured review on the parameters that show abnormalities in blood testings to a given patient when contracting the COVID-19. Table \ref{tab:anomalias_covid} contains an excerpt of the main exams that shows significant changes in lab test results for the patients analysed at this study.

There is also consistent abnormalities described in \cite{changes1}, dealing mainly with white-blood cells, platelets, C-reactive protein, AST, ALT, GGT and LDH parameters. This study concludes that some cutoffs for these tests could be applied as an alternative to rRT-PCR tests when necessary and also paving the way for automated tests using ML when more patient data becomes available for use.

In \cite{changes2}, the patients were separated using the overall gravity of the infection - that could be used as a proxy for special-care treatment as well. Main results from this study points out to significant changes comparing the patients with established reference values and within different infection gravity groups. Again, the most relevant values obtained were for white-blood cell count, LDH, C-reactive protein and others. Moreover, the article finishes stating that the virus could be related to a state of hyper-coagulation in critically-ill patients, exposing a possible interaction between COVID-19 and blood lab results. 
Knowing these facts, here we propose an extension to use the same exams data jointly with hospital outcomes to predict whether the same given patient will also need special care - effectively anticipating the use of valuable medical time and resources. We also model the number of days each patient will be on special-care using the same data.

\begin{table}[ht!]
\begin{center}
\caption{Main abnormalities found in COVID-19 patients, according to \cite{giuseppe202laboratory}.} \label{tab:anomalias_covid}
\begin{tabular}{lc}
    \hline
    Lab Exam & COVID-19 Effects \\ \hline
    Albumin & Decrease \\
    Reactive C-Protein – PCR & Increase \\
    Eritrocytes & Increase \\
    Haemoglobin & Decrease \\
    Leukocytes & Increase \\
    Neutrofils & Increase \\
    Lymphocytes & Decrease \\
    TGP-ALT & Increase \\
    TGO-AST & Increase \\
    Lactate Desidrogenase - LDH & Increase \\
    D Dimer & Increase \\
    Bilirrubin & Increase \\
    Creatinin & Increase \\
    Troponin I & Increase \\
    Procalcitonin - PCT & Increase \\
    Protrombin & Increase \\ \hline
\end{tabular}
\end{center}
\end{table}

\subsection{Machine learning procedure}
\label{mlproc}

Even without analysing the available data, one should expect from the domain of data science that three things should be present: \textbf{sparsity}, as some laboratory exams are not performed for all patients, revealing a lot of blanks (NAs) on the dataset. Also one should expect \textbf{unbalancing}, as not all patients will require special care (and in fact just a little number of them will). The last thing it is expected \textbf{non-linearity and interaction} - once every patient will have a different set of variables and the final combination and composition of them will express the outcome distinguished for each patient.

We will focus primarily in Sirio Libanês Hospital data, which includes patient outcomes and dates of admission and discharge, making it possible to analyse the number of days each patient stays under special care and associate it with the lab exams data. All data is taken for each patient, and a pre-processing is made to relate the first exam ever recorded to the patient so we preserve the time dependency which is relevant to the problem. Later exams should not constitute reliable data as they introduce temporal leaks.

In order to model properly the situation, we propose (for both targets) an two-part procedure that addresses all issues cited above.
The first part is composed by an initial exploration on data, to understand its particular shape and properties, focusing in age and blood white-cell components, as discussed in the earlier section.  After that, we explore the usage of \textit{off-the-shelf} algorithms with little to no customisation to better understand which candidate suits best - considering the baselines for each model (a coin for the special care classifier and the mean training value for the number of days target) as well the overall capacity to accept different \textit{hyperparameters} to increase the fitness of the model also considering the training time and complexity trade-offs of all algorithms as a secondary but important factor.

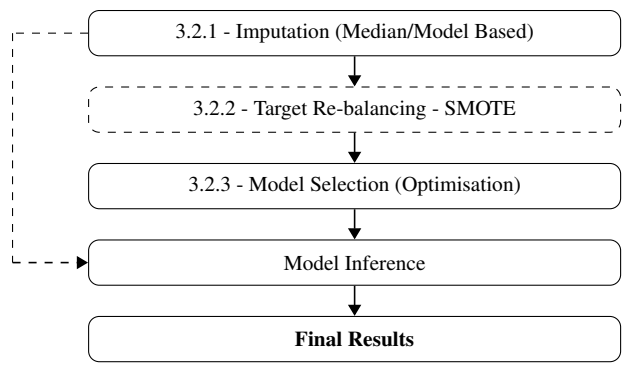
\begin{figure}
\centering
\begin{tikzpicture}[
node distance = 4mm,
  start chain = A going below,
   arr/.style = {-Triangle, semithick},
   box/.style = {rectangle, draw, rounded corners,
                 minimum width=70mm, minimum height=6mm,
                 on chain=A, join=by arr}
                        ]
\node[box](main)  {\ref{inputation} - Imputation (Median/Model Based)};
\node (point1) [draw=none,fill=none, left of=main, xshift=-4cm]{};
\node[box, dashed]  {\ref{rebalancing} - Target Re-balancing - SMOTE};
\node[box](shortcut)  {\ref{estimationopt} - Model Selection (Optimisation)};
\node[box](final)  {Model Inference};
\node (point2) [draw=none,fill=none, left of=final, xshift=-4cm]{};
\node[box]  {\textbf{Final Results}};      

\draw[black,dashed] (main.west) -- (point1.west);
\draw[black,dashed] (point1.west) -- (point2.west);
\draw[arr, black, dashed] (point2.west) --(final.west);
\end{tikzpicture}
\caption{Steps in second part for our targets. Black continuous arrows are for training phase and dash one for prediction phase. Dashed step is not applied in number of days target.}
\label{fig:steps}
\end{figure}

Once the selected class of model is chosen, we follow the procedure outlined in Figure \ref{fig:steps}, composed by data imputation, re-balancing and estimation steps. The following subsections will deal with practicalities and possible choices detailing the pros and cons for each one of the steps to pave the way to establish a detailed method that can be used on other similar situations.

\subsubsection{Imputation strategies}
\label{inputation}
To deal with data \textbf{sparsity}, we have three options with different assumptions and each one has an implication on model dynamics being dicussed in the next paragraphs. One good sparsity treatment similar to ours can be found at \cite{svd2}, a seminal article in the field.

The first one is retaining the sparsity, i.e. not applying any technique to deal with variables completion. There are two disadvantages when going this way - the first one is that most models do not handle very well with sparsity. Some of them even fail altogether during the training phase as they depend on dense matrix for parameter estimation (great part of the \textit{"classical statistical"} models fall in this category).
The second major issue is that models in general need some variance to "learn" the most relevant variables in a dataset. When a dataset is substantially sparse, some variables lose their "protagonism" and may become irrelevant even whether they are important considering the application domain. The main advantage using this approach is that data can be used \textit{as it is}, without resorting to pre-processing and cleaning.

The second major option relies on model-based variable completion, like the ones presented in \cite{svd1} and \cite{svd3}. Most of these procedures consist on Singular Value Decomposition variants, commonly used in biological and medical applications. These model-assisted matrix completion algorithms introduce interaction terms that can be very useful whether the number of patients is high enough in the dataset. The main disadvantage of this technique is the care needed to find the optimal values for each one of the hyperparameters in each one of the algorithms, in turn consuming more time and computation resources, being a barrier in its implementation for very large dataset although there is some developments in running the algorithms more efficient and parallel distributed.

The third and simpler way is just by inputting some known statistic of the sample as the default value for each variable. The most common values used for this are the mean and median (using the points with observations). Overall justification for this procedure relies on the fact that assuming that there is more healthy patients than unhealthy ones (or more patients that do not require special care), the mean and median for a sample describes a healthy population as the number of samples increase, helping models to identify abnormal values. The main disadvantage remains in the fact that some exams can be prescribed more for unhealthy (or healthy) patients therefore skewing the mean to be used as input, generating some sample bias.

In this study we choose the second and third options interchangeably in different parts of the analysis - with an special preference to use the third one, simplifying the calculations.

\subsubsection{Data re-balancing}
\label{rebalancing}
We should expect from data that not all patients require special care. Moreover, it is expected that only a few of them will. On machine learning, this type of problem is known as \textbf{unbalancing} between classes. The thing here is that having just a few samples of one specified occurrence, the model is not able to generalise well considering the few examples giving a poor specificity/sensitivity model. Here accuracy is not important because a model that gives as answer the predominant class generally will present a good value for accuracy. The Receiver Operating Characteristics (ROC) statistics can also be affected by this situation in a minor extent.

Some studies try to understand the overall effect of unbalancing on classifiers of different types. For example, \cite{unbalance1} tries to understand the overall effects in several public available datasets and even proposes changes in calculation of performance metrics that are more adequate to these situations. This is certainly an improvement over the original situation but we will make use of an other alternative which is more automated and depends less of human interaction.

Manual techniques such as under-sampling of majority class or over-sampling of minority class through bootstrapping were usually taken in consideration in the past for some studies and practical applications, with mixed results and poor reproducibility when new data arrives for model updates. To avoid this, here we will use Synthetic Minority Oversampling Technique as described by \cite{smote}, a technique to combine the minority class oversampling and synthetic example generation with majority class under-sampling, augmenting the area under the ROC curve statistics and turning the model more sensitive to minority class.

\subsubsection{Model Estimation and Optimisation}
\label{estimationopt}
When selecting models for a specific application, several aspects should be considered, the most relevant being definitely the overall "\textit{capacity}" of the algorithm - the way an specific algorithm learns about different patterns existing in data without over-fitting to it. Most algorithms regulate this capacity by the change of hyperparameters controlling different aspects.

Finding optimal hyperparameters is a matter of discussion in scientific debate since ML gained traction as an everyday tool, and as pointed by \cite{gridsearch}, still a growing field for new discoveries. 
Well-known libraries among data scientists for computational ML implement different strategies (see \cite{scikit-learn} for a good example), most of them based in grid searches of several different parameters, but there is two major disadvantages in doing this. The first and more obvious one is in the process itself requiring an high number of evaluations in the cross-validation process, being directly proportional to the number of folds. The second less apparent and more important disadvantage is the search space itself that needs to be crafted and selected (taking all relevant parameters in consideration for the problem).

While most techniques cannot deal well with the second disadvantage (crafting the search space), there is a possible improvement usually requiring less evaluations in our cross-validation procedure with its roots in optimisation and statistics. here we propose the use of Bayesian Optimisation as in \cite{bayesopt} to select model hyperparameters achieving optimal performance within the selected grid. Our procedure will be very similar to the procedure described in \cite{mlbayes}. The parameters we optimise will be discussed in the results section for the selected algorithm. 

\section{Computational Results}
\label{sec:numerical_results}

Here we present the computational results of our work, divided in three parts. First in \ref{dados} we analyse some data features of our problem, with a deep-dive in some variables already mentioned in other sections. In section \ref{preliminary} we use several algorithms with default parameters to select the best algorithm type to use together with Bayesian Optimisation considering the hyperparameters to be tuned and overall performance. In \ref{optmized} we introduce the optimised models for both targets and discuss their results.

\subsection{Data}
\label{dados}

Our dataset consists of lab exams data collected from 9633 patients from Sírio Libanês Hospital, looking for treatment in several different departments during the COVID-19 pandemic in Brazil. All patients from this list were submitted to the COVID-19 test (we included both positives and negatives) and 674 (7\%) of them required special care treatment (hospitalisation in common, semi- or intensive care units). Among the ones requiring special treatment, the mean number of days required for each patient was 1.52 days with high variation, considering its standard deviation of 6.92 days.

There is in total 165 different types of laboratory exam results (which in turn helps to understand the aforementioned \textbf{sparsity}). Considering demographics, we have age and sex available for each patient. Age will be analysed further ahead in more detail.

We first show our exploratory analysis results in Table \ref{tab:data_description} considering some statistics for the variables in dataset (for the ones with most coverage). We also show the two-sample Kolmogorov-Smirnov (KS) statistic value for each one considering special care target values as class variable, to understand overall statistical difference between distributions that can arise between classes.

\begin{center}
\begin{table*}[htbp]
\caption{Variable metrics for the ones with most coverage within dataset (146 variables omitted).}
\label{tab:data_description}
\centering
\resizebox{\columnwidth}{!}{\begin{tabular}{lrrrrrrr}
\toprule
{} &       Mean &       Std &     Min &      IQR &       Max &  Coverage (\%) &  KS Statistic \\
\midrule
Sex               &       0.46 &      0.50 &     0.0 &      1.0 &       1.0 &     100.0 &          0.00 \\
Age (\textit{years})              &      42.48 &     13.99 &    15.0 &     17.0 &      87.0 &      99.0 &          0.00 \\
MCH (\textit{pg})               &      29.16 &      2.26 &    18.0 &      2.0 &      38.0 &      18.0 &          0.17 \\
Hematocrit (\textit{\%})   &      39.61 &      5.48 &    15.0 &      6.0 &      62.0 &      18.0 &          0.00 \\
CMCH (\textit{pg})          &      33.09 &      1.23 &    27.0 &      2.0 &      37.0 &      18.0 &          0.00 \\
Erythrocytes (\textit{million/$mm^{3}$})     &       4.06 &      0.80 &     1.0 &      1.0 &       7.0 &      18.0 &          0.06 \\
Leukocytes (\textit{/$mm^{3}$})       &    6258.91 &   3541.01 &   100.0 &   3015.0 &   55110.0 &      18.0 &          0.00 \\
RDW (\textit{\%})              &      13.22 &      2.51 &    11.0 &      2.0 &      38.0 &      18.0 &          0.02 \\
Hemoglobin (\textit{g/dL})       &      12.97 &      1.99 &     5.0 &      2.0 &      21.0 &      18.0 &          0.00 \\
Platelets         &  205748.36 &  78948.08 &  7000.0 &  95000.0 &  529000.0 &      18.0 &          0.00 \\
Neutrophils (\textit{\%})   &      61.71 &     14.57 &     1.0 &     19.0 &      97.0 &      18.0 &          0.00 \\
Eosinophils (\textit{/$mm^{3}$})    &      81.96 &    112.61 &     0.0 &    100.0 &     950.0 &      18.0 &          0.00 \\
Monocites (\textit{\%})     &       9.24 &      4.49 &     0.0 &      5.0 &      43.0 &      18.0 &          0.00 \\
Eosinophils (\textit{\%})   &       1.04 &      1.72 &     0.0 &      2.0 &      14.0 &      18.0 &          0.00 \\
Lymphocytes (\textit{\%})   &      25.75 &     12.38 &     0.0 &     16.0 &      84.0 &      18.0 &          0.00 \\
Basofils (\textit{\%})     &       0.07 &      0.30 &     0.0 &      0.0 &       4.5 &      18.0 &          0.19 \\
Neutrophils (\textit{/$mm^{3}$})    &    4132.13 &   3142.68 &    20.0 &   2550.0 &   53730.0 &      18.0 &          0.00 \\
Lymphocytes  (\textit{/$mm^{3}$})     &    1463.58 &    841.17 &    20.0 &    920.0 &   14350.0 &      18.0 &          0.00 \\
Basofils   (\textit{/$mm^{3}$})      &      24.15 &     25.71 &     0.0 &     20.0 &     410.0 &      18.0 &          0.00 \\
Monocites  (\textit{/$mm^{3}$})       &     575.24 &    420.51 &    10.0 &    310.0 &    9170.0 &      18.0 &          0.00 \\
Platelet Volume   &       9.85 &      0.92 &     8.0 &      1.0 &      13.0 &      18.0 &          0.10 \\
Creatinine  (\textit{mg/dL})      &       0.51 &      0.86 &     0.0 &      1.0 &      11.0 &      16.0 &          0.00 \\
Urea    (\textit{mg/dL})          &      34.71 &     18.32 &    10.0 &     14.0 &     201.5 &      16.0 &          0.00 \\
Potassium   (\textit{mEq/L})      &       3.54 &      0.55 &     2.0 &      1.0 &       6.5 &      15.0 &          0.00 \\
Sodium   (\textit{mEq/L})         &     138.42 &      3.05 &   121.0 &      3.0 &     152.0 &      14.0 &          0.00 \\
ALT  (\textit{U/L})        &      37.26 &     38.03 &     6.0 &     25.0 &     521.0 &      13.0 &          0.00 \\
AST   (\textit{U/L})           &      35.76 &     45.41 &     9.0 &     16.0 &    1140.5 &      13.0 &          0.00 \\
DHL   (\textit{U/L})           &     488.87 &    345.04 &   201.5 &    166.0 &    8958.0 &      11.0 &          0.00 \\
\bottomrule
\end{tabular}}
\end{table*}
\end{center}

\begin{center}
\begin{figure}[htbp]
\includegraphics[width=1\columnwidth]{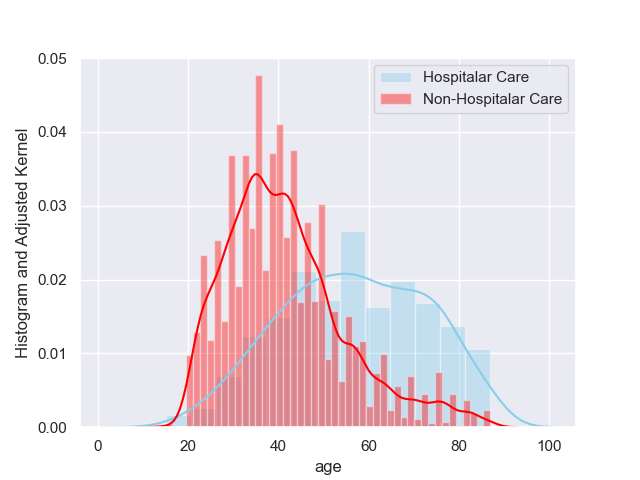}
\caption{Histogram and adjusted kernels for age, divided using the special care target.}
\label{fig:age}
\end{figure}
\end{center}

As pointed in \cite{age}, age seems to be a critical factor overall considering COVID-19 and the sample of population we are considering. Being the only continuous demographic variable, we display the class histogram for age with adjusted kernels in Figure \ref{fig:age}. We see a very distinct separation between classes arising for each one of the groups, but this pattern by itself is not substantial to make any assumptions or conclusions about our targets.

\begin{figure*}[htbp]
\centering
\includegraphics[width=1\textwidth]{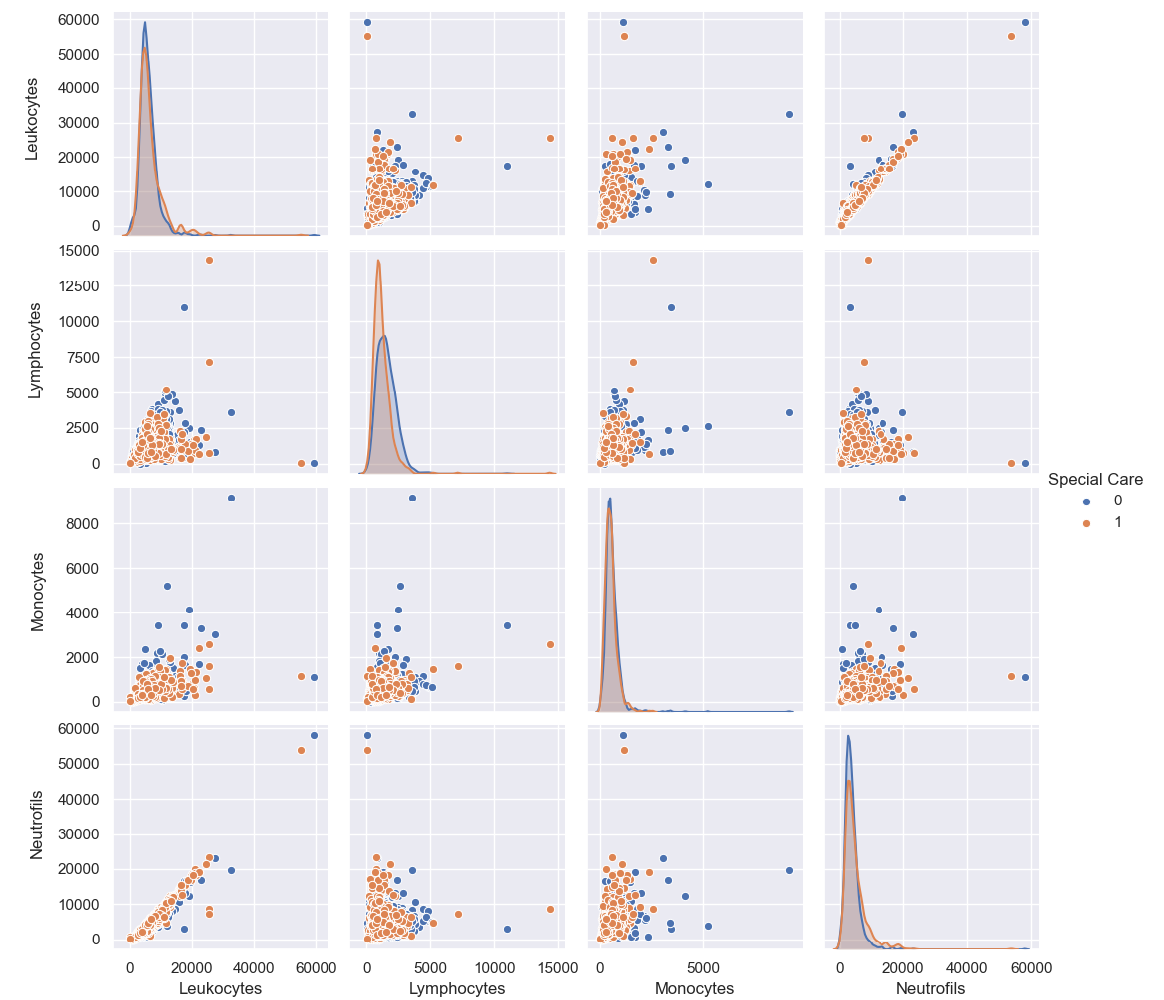}
\caption{Histogram and adjusted kernels for white-cell blood components, divided using the special care target.}
\label{fig:whitecount}
\end{figure*}

In Figure \ref{fig:whitecount}, we see the histograms and adjusted kernels for selected blood white-cell components count which superficially represent immunological responses for each one of the patients in data and also mentioned as important by other authors investigating samples coming from similar conditions, as mentioned earlier. By close inspection we see that separation for the variables considering the classes is not evident using only univariate reasoning, which again points for the necessity to use multivariate and non-linear algorithms. 

This brief analysis shows a perfect match for ML applications: We have sufficient patient data, with no identifiable univariate pattern relating to our target opening the possibilities of multivariate analysis and algorithms recognising several different types of trends and interactions (the aforementioned \textbf{non-linearity}).

\subsection{Preliminary Models}
\label{preliminary}

To begin our modelling we used several ML algorithms without any tuning on the parameters, to select the best algorithm type to be optimised later. Our tests considered Naïve Bayes, Decision Trees, AdaBoost, Support Vector Machines (SVD), Linear Descriminant Analysis (LDA), Quadratic Discriminant Analysis (QDA), Logistic Regression (and regularised ones like Ridge Regression and Least Absolute Shrinkage and Selection Operator (LASSO)), Orthogonal Matching Pursuit (OMP) and other algorithms based on ensembles of trees, like Extra Trees \cite{extratrees}, Random Forests \cite{RF}, xgBoost \cite{xgboost}, and LightGBM \cite{lightgbm}.
All results were obtained using Python 3.7 as programming language. To obtain the following results data was treated as-is, i.e. without any treatment or imputation strategies being used.

Model type selection for further optimisation should consider three key practical aspects, with more importance to the first two. The first one is predictive power - we want an algorithm that predicts well and do not overfit to our data while capturing the multivariate effects that we expect. The second aspect involves the number of hyperparameters available to tune in the model. The more parameters, more opportunities we have to improve our algorithm in predictive power, while keeping the generalisation capacity. The third one is time to perform training - albeit less important may generate hassle when datasets are large enough, being considered even in our context because the subsequent optimisation we perform requires several full training passes in data.
Table \ref{tab:prelim_internacao} presents results considering algorithms for the special care target and all relevant metrics. The baseline for this models is a coin with ROC AUC value of $0.5$ .
Table \ref{tab:prelim_dias} presents results and relevant metrics for number of days under special care target. The baseline here is the mean value of the training set.

\begin{table*}[htbp]
\caption{Results from preliminary models on special care target (Top 10 of all models tested). Chosen algorithm for optimisation is highlighted.}
\label{tab:prelim_internacao}

\centering
    	
\begin{tabular}{lrrrr}
\toprule
{} &  Balanced Accuracy &  ROC AUC &  F1 Score &  Time Taken (s) \\
Model                         &                    &          &           &             \\
\midrule
Bernoulli Naïve Bayes                   &               0.90 &     0.90 &      0.92 &        0.14 \\
QDA &               0.88 &     0.88 &      0.91 &        0.22 \\
Gausssian Naïve Bayes                    &               0.85 &     0.85 &      0.95 &        0.15 \\
\textbf{xgBoost}                 &               0.85 &     0.85 &      0.96 &        1.31 \\
LightGBM                &               0.82 &     0.82 &      0.96 &        0.47 \\
AdaBoost            &               0.82 &     0.82 &      0.96 &        0.92 \\
SVC                           &               0.81 &     0.81 &      0.95 &        2.52 \\
Random Forest        &               0.81 &     0.81 &      0.96 &        1.14 \\
Baging              &               0.80 &     0.80 &      0.96 &        0.78 \\
Decision Tree        &               0.80 &     0.80 &      0.96 &        0.23 \\
\bottomrule
\end{tabular}
\end{table*}

\begin{table*}[htbp]
\caption{Results from models on number of days of special care needed (Top 10 of all models tested). Chosen algorithm for optimization is highlighted.}
\label{tab:prelim_dias}
\centering
\begin{tabular}{lrrr}
\toprule
{} &                  R-Squared &            RMSE &  Time Taken (s) \\
Model                         &                            &                 &             \\
\midrule
\textbf{xgBoost}                  &                       0.70 &            2.15 &        1.28 \\
Vanilla Gradient Boosting     &                       0.68 &            2.22 &        1.98 \\
Random Forest         &                       0.66 &            2.31 &        7.67 \\
Bagging           &                       0.64 &            2.38 &        0.92 \\
LightGBM                  &                       0.60 &            2.49 &        0.36 \\
Extra Trees            &                       0.60 &            2.50 &        9.65 \\
Histogram Gradient Boosting &                       0.60 &            2.52 &        4.75 \\
Huber Regression         &                       0.45 &            2.94 &        1.70 \\
LinearSVR                     &                       0.44 &            2.96 &        3.14 \\
Decision Tree         &                       0.43 &            2.99 &        0.22 \\
\bottomrule
\end{tabular}
\end{table*}

The final selected algorithm is xgBoost for both targets. The main rationale for this are the aforementioned characteristics: high predictive power, hyperparameter tuning and overall training time. We also could select LightGBM interchangeably as the results were very close (and the algorithms are similar as well), besides it was faster. Between the two algorithms, our previous experience with xgBoost motivated us to choose it. It stands out algorithms like Naïve Bayes have almost no hyperparameters to tune and were unconsidered even performing very well in the preliminary analysis.

\subsection{Optimised Models}
\label{optmized}

Having selected the final algorithm type to use, we must define which hyperparameters to use in Bayesian Optimisation and also which strategy to deal with sparsity and unbalancing. Table \ref{tab:parameters} shows all parameters being considered in the Bayesian Optimisation and its respective intervals and descriptions. All optimisation is performed using Ax \cite{ax}, a platform created inside Facebook that streamlines all optimisation process and makes possible the use of integer hyperparameters, which is not available in other solvers.

\begin{table*}[htbp]
\caption{Parameter grid and intervals used in Bayesian Optimisation procedure.}
\label{tab:parameters}

\centering
    	
\begin{tabular}{lll}
\toprule
{} & Interval & Description \\
\midrule
eta     &  $[0.01, 1]$  & Learning rate (shrinkage applied in weights calculation) \\
gamma      &  $[0, 100]$ & Minimum loss reduction to split a node in tree \\
max\_depth &  $[1,9]$ & Maximum depth of each tree in training process \\
subsample  &  $[0.5, 1]$ & Number of features used to train a tree \\
lambda  &  $[1, 100]$ &  $L_2$ regularization term using in training \\
alpha  &  $[0, 100]$ &  $L_1$ regularization term using in training \\
n\_estimators  &  $[10, 200]$ & Total number of trees \\
\bottomrule
\end{tabular}
\end{table*}

\begin{figure}[htbp]
\centering
\includegraphics[width=1\columnwidth]{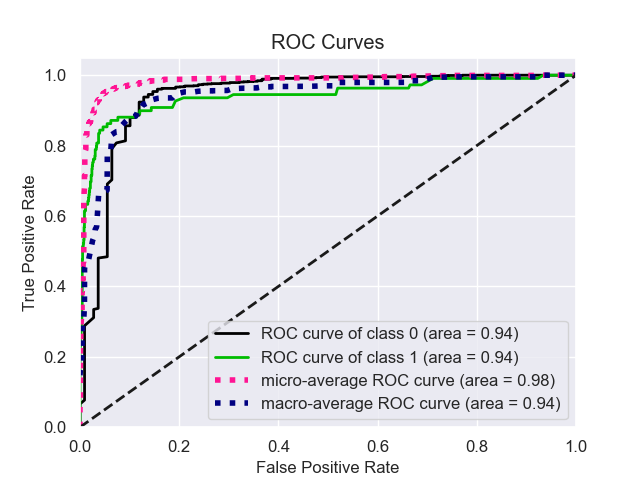}
\caption{ROC Curve for special care target, both classes.}
\label{fig:roc_auc}
\end{figure}

\begin{figure}[htbp]
\centering
\includegraphics[width=1\columnwidth]{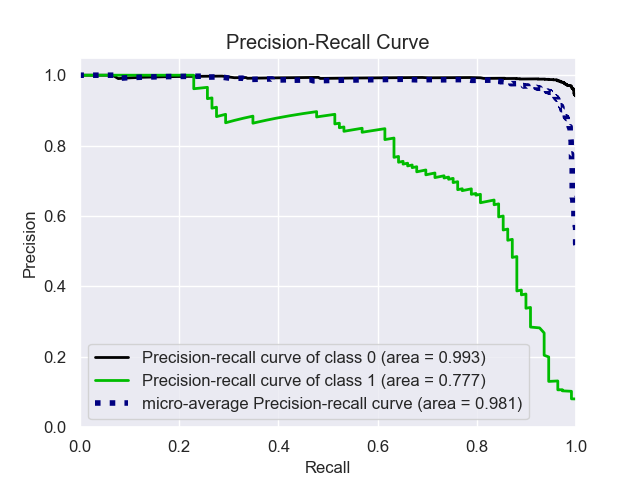}
\caption{Precision-Recall Curves for special care target, both classes.}
\label{fig:pr}
\end{figure}

For a classification model to be useful, we need not only to look at Receiver Operating Characteristic (ROC) curves but to look as well for Precision-Recall (P/R) curves, which can be a total different format considering the variable distribution. Figure \ref{fig:roc_auc} summarizes the ROC curve and Figure \ref{fig:pr} summarizes the P/R curve. In our tests, using median as imputer gave us the best results overall for the special care target.

We see as well that our optimization improved the ROC statistic by selecting a new set of hyperparameters different than the defaults. By doing that, we guarantee that we have the best model while keeping model generalisation capabilities.

From a hospital perspective, False Positives (the abscissa from our ROC plot) constitutes most of the lost resources. They are patients that do not need any special care, but the model indicates opposite and we should keep them on a minimum level. We see by close inspection of the curves that this is satisfied, and the model is indeed useful to classify patients using blood-test samples. Summarising, at the best threshold value for cutoff we got 0.94 for ROC AUC, and 0.77 for P/R AUC.

\begin{figure}[htbp]
\centering
\includegraphics[width=0.8\columnwidth]{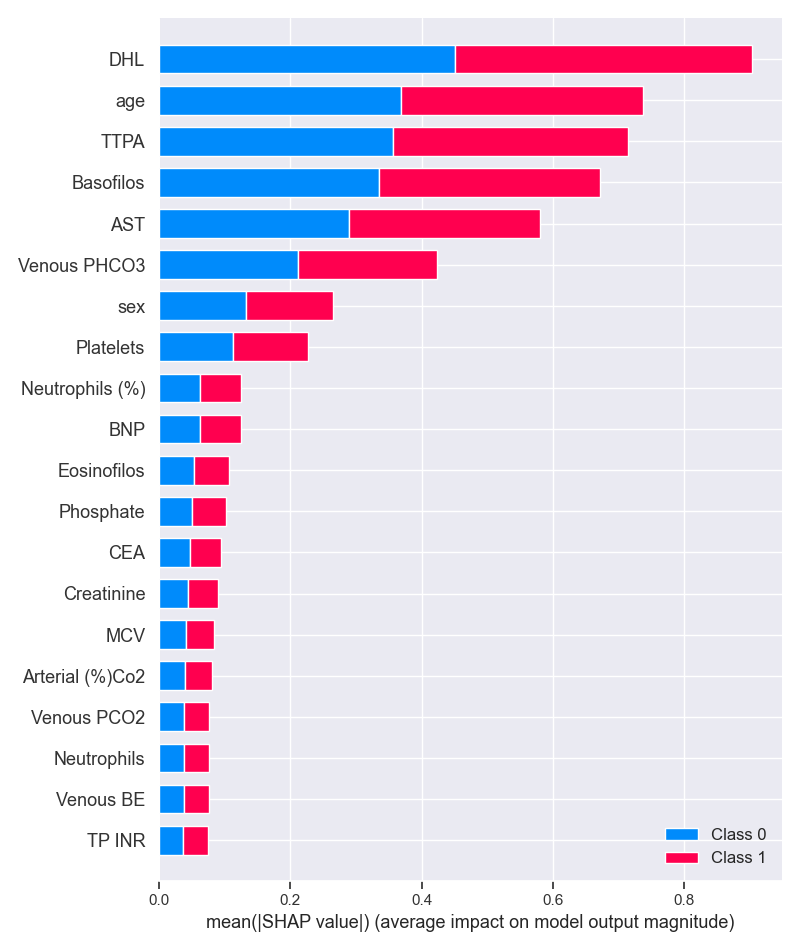}
\caption{Variable importance plot for special care target, both classes.}
\label{fig:varimp}
\end{figure}

Moreover, as we used ensembles of trees to make predictions, one thing that arises naturally is a variable importance plot. To obtain this plot we used Shap \cite{shap}, which creates this plot using a game-theoretical approach to calculate the variable importance for row and data levels. In Figure \ref{fig:varimp}, we see that some of the variables presented as important (mentioned in Section \ref{sec:method}) in \cite{changes1} and \cite{changes2} are indeed some of the most relevant in our model as well, going in line with the expectations (This plot should not be seen as indicating any direct causal relationships as our data is not experimental, but observational).

Results for the days under special care were similar in performance achievements. Table \ref{tab:days_tab} summarises the findings and compare them with the baseline for this model, which is the mean value of days spent in special care for the training set. Best results were obtained using no imputer at all (using model based input gave us the worst results in comparison), defying some preconceptions we had from the start. This effect is explained in \cite{elsl}: adding variables to boosted or bagged regressors can make the model worse. By using imputers, we forced the model to be non-sparse, giving protagonism to all variables at once, amplifying this condition. The condition for classifiers is the opposite: adding variables to boosted or bagged models always increases the performance (but the improvement could be marginal).

\begin{figure}[htbp]
\centering
\includegraphics[width=1\columnwidth]{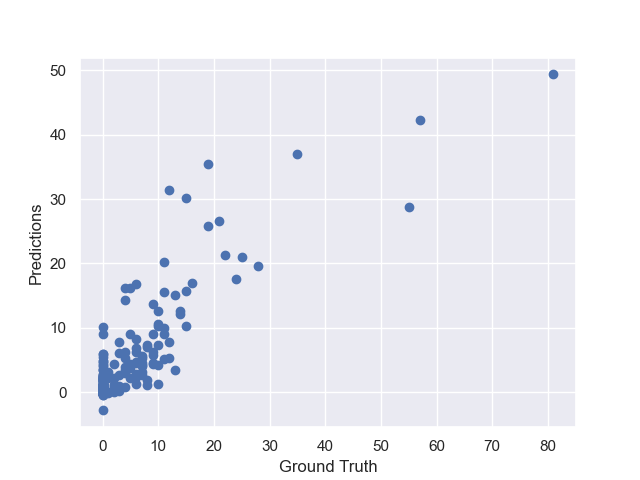}
\caption{Scatterplot for days under special care target.}
\label{fig:scatter}
\end{figure}

Although our model is capable of doing good predictions as guaranteed by statistical tests, in Figure \ref{fig:scatter} we see a tendency to \textit{overshoot} and \textit{undershoot} the results caused by the very nature of model (splits in trees have a very poor tendency in addressing extreme situations as the capacity to extrapolate wanes as we go to the ends of our interval) . More discussion on model improvement is placed in Section \ref{sec:limit}.

\begin{table}[htbp]
\caption{Results for days under special care target, baseline and percentual improvement over baseline.}
\label{tab:days_tab}
\centering
    	
\begin{tabular}{lllr}
\toprule
{} & Model & Baseline &  Improvement (\%) \\
\midrule
RMSE &  1.87 &     3.96 &            77.78 \\
MAE  &  0.41 &     1.27 &            67.96 \\
R-Squared   &  0.78 &     0.00 &       -- \\
\bottomrule
\end{tabular}

\end{table}

\section{Limitations and Possible Extensions}
\label{sec:limit}
So far with our classifier model we were dealing with 0/1 outcomes only; We were just trying to detect whether any patient will require special care or not. But, if we want to order our patients from the one with less \textit{"risk"} to the one with more \textit{"risk"}? (Risk here being associated with an well established and concise probability measure ranging from 0 to 1). As a matter of fact, the algorithm used to learn our special care from data is not well suited for this specific task. In \cite{calibration}, this effect is described as the algorithms having difficulties making predictions near the frontiers of the [0,1] interval because the variance of the base trees drives the result away from the borders in a way to minimise the overall cost function. To diagnose this problem, one can calculate the overall Brier score \cite{brier} for a given model or make an calibration plot. To solve this issue we could apply Platt's method \cite{platt}, which essentially adjusts an Logistic Regression on a different fold during model training phase, or applying an Isotonic Regression \cite{isotonic}, again on a different fold during model training, but more data for patients is required to perform that in a meaningful way.

To deal with negative predictions arising in the number of days under special care target, we must understand first that the model used for making the predictions is not restricted in any form about the prediction interval itself. All of its predictions lie within the real line ${\rm I\!R}$, but we know that our values are at least limited by 0. A recent way to deal with this is emerging on disciplines such as Finance and Banking, exposed in \cite{grabit} where ensembles of trees are trained to perform Tobit regression. The overall maturity for the packages is increasing fast, posing as an very interesting development as ensembles of trees have very high predictive power in general and several hyperparameters that can be optimised using Bayesian Optimisation in the same process.

To deal with \textit{overshooting} and \textit{undershooting} for our number of days under special care target, there is several possibilities arising from traditional statistics worth exploring like the Zero Inflated Negative Binomial (ZINB) models \cite{zinb} in which the target distribution is composed of a very high proportion of zeroes, like our target. The result for this type of models usually consists of a probability attached with a counter, probability measuring the overall chance of a given patient needing special care and the counter giving the number of days the same patient will spend under such care. The major drawback for this from the model is the predictive power (especially for the probability part), where common packages use only linear terms (which in turn introduce needs on data pre-processing, like multicollinearity removal or variance inflation factors analysis) and no ensembles to make predictions. An viable but not tested alternative could be mixing two "worlds" together, trying different sets of variables on the dataset guided by Bayesian Optimisation and then applying an ZINB model for each one, averaging the results. The counter model in this situation is discrete, also solving the issue with non-integer predictions.

\section{Final remarks}
\label{sec:conclusion}
The growing necessity to predict the needs of hospital resources guided the exploration of novel methods to help create and plan policies accessible for everyone and more than ever the COVID-19 pandemic is pushing health systems to the limit. Having this in mind, we developed an analytical approach based on mathematical model and algorithms using the most recent techniques available in the fields of statistics and machine learning using public data available.

We consider have achieved interesting results in this study. The estimated 0.94 area under ROC Curve combined with 0.77 P/R statistic proves that the analytical approach can indeed be used in a decision system for hospitals, governments and health providers alike to guide their resource allocation with minimal requirements as we use exam data that is available and mostly affordable. The target for number of days under special care certainly needs refinement but is also in good shape in our view. Other interesting results are also in line with other studies done by researchers all around the world.

Besides the proposal of a decision system, our main objective is to spread the use of these tools to provide systems that can be used not only for COVID-19, but for other diseases - increasing overall planning maturity and patient service levels.

\vspace{8pt}
\noindent\textbf{Acknowledgements:}
    The authors would like to thank the anonymous referees for their useful comments. This project partially benefited from the CNPq grant (process number: 400868/2016-4).

\bibliographystyle{spmpsci}      
\bibliography{main}   

\begin{thebibliography}{10}
\providecommand{\url}[1]{{#1}}
\providecommand{\urlprefix}{URL }
\expandafter\ifx\csname urlstyle\endcsname\relax
  \providecommand{\doi}[1]{DOI~\discretionary{}{}{}#1}\else
  \providecommand{\doi}{DOI~\discretionary{}{}{}\begingroup
  \urlstyle{rm}\Url}\fi

\bibitem{MLHC}
Machine learning for healthcare (2020).
\newblock \urlprefix\url{https://www.mlforhc.org/}

\bibitem{blood}
Alsheref, F.K., Gomaa, W.H.: Blood diseases detection using classical machine
  learning algorithms.
\newblock International Journal of Advanced Computer Science and Applications
  \textbf{10}(9) (2019)

\bibitem{ax}
Bakshy, E., Dworkin, L., et~al.: Ae: A domain-agnostic platform for adaptive
  experimentation.
\newblock NIPS'18: Proceedings of the 31th International Conference on Neural
  Information Processing Systems  (2018)

\bibitem{bacteremia}
Beeler, C., Dbeibo, L., et~al.: Assessing patient risk of central
  line-associated bacteremia via machine learning.
\newblock American Journal Infect Control \textbf{46}(9), 986--991 (2018)

\bibitem{bertsimas}
Bertsimas, D., OHair, A.K., Pulleyblank, W.R.: The Analytics Edge.
\newblock Dynamic Ideas LLC, Belmont MA (2015)

\bibitem{age}
Bonanad, C., García-Blas, S., et~al.: The effect of age on mortality in
  patients with covid-19: A meta-analysis with 611.583 subjects.
\newblock Journal of the American Medical Directors Association \textbf{21},
  915--918 (2020)

\bibitem{RF}
Breiman, L.: Random forests.
\newblock Machine Learning \textbf{45}, 5--32 (2001)

\bibitem{brier}
Brier, G.W.: Grabit: Gradient tree-boosted tobit models for default prediction.
\newblock Monthly Weather Review \textbf{78}(1) (1950)

\bibitem{covid1}
Brinati, D., Campagner, A., et~al.: Detection of covid-19 infection from
  routine blood exams with machine learning: A feasibility study.
\newblock Journal of Medical Systems \textbf{44}(8) (2020)

\bibitem{xgboost}
Chen, T., Guestrin, C.: Xgboost: A scalable tree boosting system.
\newblock KDD  (2016)

\bibitem{ebola}
Colubri, A., Silver, T., et~al.: Transforming clinical data into actionable
  prognosis models: Machine-learning framework and field-deployable app to
  predict outcome of ebola patients.
\newblock PLOS Neglected Tropical Diseases \textbf{10}(3) (2016)

\bibitem{ibero}
Dutta, S., Bandyopadhyay, S.K.: Machine learning approach for confirmation of
  covid-19 cases: positive, negative, death and release.
\newblock Iberoamerican Journal of Medicine \textbf{03}, 172--177 (2020)

\bibitem{newone}
Elaziz, M.A., Hosny, K.M., et~al.: New machine learning method for image-based
  diagnosis of covid-19.
\newblock PLOS ONE  (2020)

\bibitem{data}
FAPESP: Covid-19 data sharing - br (2020).
\newblock
  \urlprefix\url{https://repositoriodatasharingfapesp.uspdigital.usp.br/}

\bibitem{survey}
Fatima, M., Pasha, M.: Survey of machine learning algorithms for disease
  diagnostic.
\newblock Journal of Intelligent Learning Systems and Applications
  \textbf{9}(1) (2017)

\bibitem{changes1}
Ferrari, D., Motta, A., et~al.: Routine blood tests as a potential diagnostic
  tool for covid-19.
\newblock Clinical Chemistry and Laboratory Medicine  (2020)

\bibitem{gridsearch}
Feurer, M., Hutter, F.: Hyperparameter Optimization. In: Automated Machine
  Learning, The Springer Series on Challenges in Machine Learning.
\newblock Springer, Cham. (2019)

\bibitem{extratrees}
Geurts, P., Ernst, D., Wehenkel, L.: Extremely randomized trees.
\newblock Machine Learning \textbf{63}, 3--42 (2006)

\bibitem{hemato}
Gunčar, G., Kukar, M., et~al.: An application of machine learning to
  haematological diagnosis.
\newblock Scientific Reports \textbf{8}(411) (2018)

\bibitem{zinb}
Hall, D.B.: Zero-inflated poisson and binomial regression with random effects:
  a case study.
\newblock Biometrics \textbf{56}(4), 1030--1039 (2000)

\bibitem{elsl}
Hastie, T., Tibshirani, R., Friedman, J.: The Elements of Statistical Learning.
\newblock Springer-Verlag, New York (2009)

\bibitem{decision}
Jain, V., Chatterjee, J.M.: Machine Learning with Health Care Perspective.
\newblock Springer International Publishing (2020)

\bibitem{lightgbm}
Ke, G., Meng, Q., et~al.: Lightgbm: A highly efficient gradient boosting
  decision tree.
\newblock Conference on Neural Information Processing Systems  (2017)

\bibitem{text}
Khanday, A.M.U.D., Rabani, S.T., et~al.: Machine learning based approaches for
  detecting covid-19 using clinical text data.
\newblock International Journal of Information Technology \textbf{12}, 731--739
  (2020)

\bibitem{svd1}
Kumar, B.: A novel latent factor model for recommender system.
\newblock Journal of Information Systems and Technology Management
  \textbf{13}(3) (2016)

\bibitem{soliton}
Lalmuanawma, S., Hussain, J., Chhakchhuak, L.: Applications of machine learning
  and artificial intelligence for covid-19 (sars-cov-2) pandemic: A review.
\newblock Chaos, Solitons \& Fractals \textbf{139} (2020)

\bibitem{giuseppe202laboratory}
Lippi, G., Plebani, M.: Laboratory abnormalities in patients with covid-2019
  infection.
\newblock Clinical Chemistry and Laboratory Medicine (0) (2020)

\bibitem{oncology}
Liu, M., Oxnard, G., et~al.: Sensitive and specific multi-cancer detection and
  localization using methylation signatures in cell-free dna.
\newblock Annals of Oncology \textbf{31}(6) (2020)

\bibitem{shap}
Lundberg, S.M., Lee, S.I.: A unified approach to interpreting model
  predictions.
\newblock In: I.~Guyon, U.V. Luxburg, S.~Bengio, H.~Wallach, R.~Fergus,
  S.~Vishwanathan, R.~Garnett (eds.) Advances in Neural Information Processing
  Systems 30, pp. 4765--4774. Curran Associates, Inc. (2017)

\bibitem{framingham}
Mahmood, S.S., Levy, D., et~al.: The framingham heart study and the
  epidemiology of cardiovascular diseases: A historical perspective.
\newblock Lancet (Mar 15 383(9921): 999–1008) (2014)

\bibitem{svd2}
Mazumder, R., Hastie, T., Tibshirani, R.: Spectral regularization algorithms
  for learning large incomplete matrices.
\newblock Journal of Machine Learning Research \textbf{11}, 2287--2322 (2010)

\bibitem{bayesopt}
Mockus, J.: Application of bayesian approach to numerical methods of global and
  stochastic optimization.
\newblock Journal of Global Optimization \textbf{4}, 347--365 (1994)

\bibitem{smote}
Nguyen, M.H.: Smote: synthetic minority over-sampling technique.
\newblock Journal of Artificial Intelligence Research pp. 321--357 (2002)

\bibitem{unbalance1}
Nguyen, M.H.: Impacts of unbalanced test data on the evaluation of
  classification methods.
\newblock International Journal of Advanced Computer Science and Applications
  \textbf{10}(3) (2019)

\bibitem{calibration}
Niculescu-Mizil, A., Caruana, R.: Predicting good probabilities with supervised
  learning.
\newblock Proceedings of the 22 nd International Conference on Machine Learning
   (2005)

\bibitem{scikit-learn}
Pedregosa, F., Varoquaux, G., et~al.: Scikit-learn: Machine learning in
  {P}ython.
\newblock Journal of Machine Learning Research \textbf{12}, 2825--2830 (2011)

\bibitem{collab}
Peiffer-Smadja, N., Maatoug, R., et~al.: Machine learning for covid-19 needs
  global collaboration and data-sharing.
\newblock Nature Machine Intelligence \textbf{2}, 293--294 (2020)

\bibitem{virusR0}
Sanche, S., Lin, Y.T., et~al.: High contagiousness and rapid spread of severe
  acute respiratory syndrome coronavirus 2.
\newblock Emerging Infectious Diseases \textbf{26}(7) (2020)

\bibitem{grabit}
Sigrist, F., Hirnschall, C.: Grabit: Gradient tree-boosted tobit models for
  default prediction.
\newblock Journal of Banking \& Finance \textbf{102}, 177--192 (2019)

\bibitem{platt}
Smola, A.J., Bartlett, P.: Advances in Large-Margin Classifiers.
\newblock MIT Press, Cambridge MA (2000)

\bibitem{mlbayes}
Snoek, J., Larochelle, H., Adams, R.P.: Practical bayesian optimization of
  machine learning algorithms.
\newblock NIPS'12: Proceedings of the 25th International Conference on Neural
  Information Processing Systems \textbf{2}, 2951--2959 (2012)

\bibitem{svd3}
Troyanskaya, O., Cantor, M., et~al.: Missing value estimation methods for dna
  microarrays.
\newblock Bioinformatics \textbf{17} (2001)

\bibitem{fattyliver}
Wu, C.C., Yeh, W.C., et~al.: Prediction of fatty liver disease using machine
  learning algorithms.
\newblock Computer Methods and Programs in Biomedicine \textbf{170}, 23--29
  (2019)

\bibitem{isotonic}
Wu, W.B., Woodroofe, M., et~al.: Isotonic regression: Another look at the
  changepoint problem.
\newblock Biometrika \textbf{88}(3), 793--804 (2001)

\bibitem{changes2}
Yuan, X., Huang, W., Ye, B., et~al.: Changes of hematological and immunological
  parameters in covid-19 patients.
\newblock International Journal of Hematology  (2020)

\end{thebibliography}
\end{document}